\documentclass[graybox]{svmult}

\usepackage{type1cm}        % activate if the above 3 fonts are
\usepackage{makeidx}         % allows index generation
\usepackage{graphicx}        % standard LaTeX graphics tool
\usepackage{multicol}        % used for the two-column index
\usepackage{multirow}        % used for the two-column index
\usepackage[bottom]{footmisc}% places footnotes at page bottom

\usepackage{framed}

\usepackage{newtxtext}       % 
\usepackage{newtxmath}       % selects Times Roman as basic font

\usepackage{color}
\usepackage{subfig}

\usepackage{array}
\newcolumntype{L}[1]{>{\raggedright\let\newline\\\arraybackslash\hspace{0pt}}m{#1}}
\newcolumntype{C}[1]{>{\centering\let\newline\\\arraybackslash\hspace{0pt}}m{#1}}
\newcolumntype{R}[1]{>{\raggedleft\let\newline\\\arraybackslash\hspace{0pt}}m{#1}}

\makeindex             

\begin{document}

\title*{Association rules over time}

\author{Iztok Fister Jr. and Iztok Fister}

\institute{Iztok Fister Jr. \at University of Maribor, Faculty of Electrical Engineering and Computer Science, Koro\v{s}ka cesta 46, Slovenia, \email{iztok.fister1@um.si}
\and Iztok Fister \at University of Maribor, Faculty of Electrical Engineering and Computer Science, Koro\v{s}ka cesta 46, Slovenia, \email{iztok.fister@um.si}}

\maketitle

\abstract{Decisions made nowadays by Artificial Intelligence powered systems are usually hard for users to understand. One of the more important issues faced by developers is exposed as how to create more explainable Machine Learning models. In line with this, more explainable techniques need to be developed, where visual explanation also plays a  more important role. This technique could also be applied successfully for explaining the results of Association Rule Mining.This Chapter focuses on two issues: (1) How to discover the relevant association rules, and (2) How to express relations between more attributes visually. For the solution of the first issue, the proposed method uses  Differential Evolution, while Sankey diagrams are adopted to solve the second one. This method was applied to a transaction database containing data generated by an amateur cyclist in past seasons, using a mobile device worn during the realization of training sessions that is divided into four time periods. The results of visualization showed that a trend in improving performance of an athlete can be indicated by changing the attributes appearing in the selected association rules in different time periods.}

\begin{framed}
Citation details: Fister Jr. I., Fister I. Association rules over time. In: Frontier advances of nature inspired industrial optimization, Springer Tracts in Nature-Inspired Computing (STNIC), 2021.
\end{framed}

\section{Introduction}
Nowadays, Artificial Intelligence (AI) powered systems are sophisticated to such an extent that they actually do not need human intervention for their design and deployment. On the other hand, their decisions ultimately start to affect human lives concerning medicine, law, economy, and defense~\cite{barredo2020explainable}. As a result, these sophisticated decisions demand some understanding, on which basis these are furnished by AI methods~\cite{goodman2016regulations}.

The problem is not crucial considering older AI-powered systems, because these are easy to interpret. Actually, they act as black-boxes, and, as such, are easily understandable by the users. Obviously, the user wants to understand the mechanisms, by which the Machine Learning (ML) models work~\cite{lipton2018mythos}. When the AI-powered system is transparent, this means that decisions made by such ML models are justifiable, legitimate, and allow obtaining detailed explanations of their behavior~\cite{gunning2017explainable}.

Explainable AI (EAI)~\cite{gunning2017explainable} has emerged in order to avoid the limitations of the current AI systems. Primarily, two issues were placed before the new domain as follows~\cite{barredo2020explainable}: (1) To create more explainable ML models by maintaining a high level of learning performance, and (2) To enable humans to understand, trust, and manage the emerging platform of AI partners effectively. In the sense of the first issue, more post-hoc explainable techniques have been developed, where visualization explanation plays an important role. Post-hoc means that the explainable techniques are applied after obtaining the results of specific ML methods, in order to discover any additional knowledge hidden in data, and, thus, help the user to understand the results properly.

Association Rule Mining (ARM) is a well known ML method that normally generates a huge amount of association rules, from which it is hard to make the proper decisions easily. An additional problem is presented by the complex form of the results, which are represented as an implication $X\Rightarrow Y$, meaning ''if antecedent $X$ is \textit{true} then its consequent $Y$ is also \textit{true}''. As a result, there we are confronted with two problems: (1) How to discover the relevant association rules in a huge archive, and (2) How to express the relations between attributes in those implication rules visually in a form that is understandable and interpretable by users.

Interestingly, the visualization of association rules has rarely been treated in literature. Indeed, the papers which referred to these topics can be summarized in the following review: The authors in~\cite{wong1999visualizing} presented a design that is able to handle hundreds of multiple antecedent association rules in a three-dimensional display with minimum human interaction, low occlusion percentage, and no screen swapping. The authors in~\cite{hofmann2000visualizing} show that the use of Mosaic plots and their variant, called Double Decker plots, can be applied for visualizing association rules. Ong et al.~\cite{ong2002crystalclear} prototyped the two visualizations, called grid view and tree view, for visualizing the association rules in their application called CrystalClear. Appice and Buono~\cite{appice2005analyzing} presented a graph-based visualization that supports data miners in the analysis of multi-level spatial association rules, while Herawan et al.~\cite{herawan2009smarviz} proposed an approach for visualizing soft maximal association rules. A very interesting interactive visualization technique, which lets the user navigate through a hierarchy of association rule groups is presented in paper~\cite{hahsler2011visualizing}. The authors in paper~\cite{jiang2008finite} explored the Hasse diagrams for the visualization of Boolean association rules. Fister et al.~\cite{Fister2019Discovering} proposed a method for identifying dependencies among mined association rules based on population-based metaheuristics~\cite{Fister2013BriefReview,gupta2018evolutionary,gupta2020plant} and complex networks, while the paper of Fister Jr. et al.~\cite{fisterjr2020information} looked for a visualization method capable of telling stories based on mined association rules. However, there are also generic tools, like the CloseViz~\cite{carmichael2010closeviz} and the SPMF open-source data mining library Version 2~\cite{fournier2016spmf}, specialized primarily in pattern mining, offering visual implementation of discovered data mined by ML algorithms that could also be used for visualization of ARM. 

This Chapter proposes monitoring the performance of an athlete involved in sport training sessions during the past seasons (i.e., years). Thus, these seasons are divided into four time periods, where each of the periods captures a quarter of the data. The performances of the athlete are measured using real data obtained by mobile devices worn by the athlete during the training sessions. The performance data (e.g., duration, average heart rate, etc.) are stored as attributes into a transaction database, from which the knowledge hidden in these is discovered using the ARM methods. These methods normally produce a huge amount of data stored in the so-called ARM archives. Data stored in these archives are also distinguished according to specific time periods. These periods, indeed, present the flow of performance of the athlete, while transitions from one period to another highlight how the form of the particular athlete progressed. 

In order to illustrate information from ARM archives in an understandable and interpretable way to users, Sankey diagrams are selected, that enable multivariate environment and historical data~\cite{lehrman2018visualizing}. The method was applied to a transaction database created by an amateur athlete in the past seasons over the duration of four and a half years. The results of visualization allow a sport trainer to analyze how the performance of the athlete in sport training improved during the seasons, on the one hand, and which attributes in which of the time periods affected these the most, on the other.

The structure of the remainder of the chapter is as follows: Section~\ref{sec:2} deals with the basic information needed for understanding the sections that follow. A description of a method for creating the Sankey diagrams from the ARM archive is the subject of Section~\ref{sec:3}. The experiments and the results are illustrated in Section~\ref{sec:4}. Chapter concludes with Section~\ref{sec:5}, that summarizes the performed and outlines the future work.

\section{Basic information} \label{sec:2}
This section is devoted to explaining the concepts necessary for a reader to understand the subjects that follow. In line with this, the following concepts are highlighted:
\begin{itemize}
    \item ARM,
    \item DE,
    \item Sankey diagram,
    \item formal definition of objectives.
\end{itemize}
In the remainder of the section, the mentioned subjects are discussed in detail.

\subsection{Association Rule Mining}
ARM can be defined formally as follows: Let us assume a set of objects $O=\{o_1,\ldots,o_M\}$ and transaction dataset $T_D=\{T\}$ are given, where each transaction $T$ is a subset of objects $T\subseteq O$. Then, an association rule is defined as an implication:
\begin{equation}
    X\Rightarrow Y,
\end{equation}
where $X\subset O$, $Y\subset O$, and $X\cap Y=\emptyset$. In order to estimate the quality of the mined association rule, two measures are defined: support and confidence. The support is defined as:
\begin{equation}
    \mathit{supp}(X\Rightarrow Y)=\frac{\#(X\cup Y)}{N},
\end{equation}
whereas confidence as:
\begin{equation}
    \mathit{conf}(X\Rightarrow Y)=\frac{\#(X\cup Y)}{\#(X)},
\end{equation}
where the function $\#(.)$ calculates the number of repetitions of a particular rule within $T_D$, and $N$ is the total number of transactions in $T_D$. Let us emphasize that two additional variables are defined, i.e. minimum support $S_{min}$ and minimum confidence $C_{min}$. These variables denote a threshold value limiting the particular association rule with lower confidence and support from being taken into consideration.

\subsection{Differential Evolution}
DE is appropriate for solving continuous, as well as, discrete optimization problems. Many DE variants have been proposed since the algorithm's birth in 1995. The original DE algorithm manages a population of real-valued vectors in the form:
\begin{equation}
    \mathbf{x}^{(t)}=(x^{(t)}_{i,1},\ldots,x^{(t)}_{i,d}),
\end{equation}
where $d$ denotes the dimension of the problem and $t$ is a generation counter. These vectors undergo operations of operators, such as mutation, crossover, and selection. In the basic mutation, two solutions are selected randomly and their scaled difference is added to the third solution, as follows: 
\begin{equation}
\vspace{-2mm}
\label{eq:de_mut}
 \mathbf{u}_{i}^{(t)}=\mathbf{x}_{r0}^{(t)}+F \cdotp (\mathbf{x}_{r1}^{(t)}-\mathbf{x}_{r2}^{(t)}),\quad\textnormal{for}\ i=1, \ldots, \mathit{NP},
%\vspace{-2mm}
\end{equation}
\noindent where $F \in [0.1,1.0]$ denotes the scaling factor that scales the rate of modification, $\mathit{NP}$ represents the population size and $r0,\ r1,\ r2$ are randomly selected values in the interval $1, \ldots, \mathit{NP}$. Note that the proposed interval of values for parameter $F$ was enforced in the DE community.

DE employs a binomial (denoted as 'bin') or exponential (denoted as 'exp') crossover. The trial vector is built from parameter values copied from either the mutant vector generated by Eq.~(\ref{eq:de_mut}) or parent at the same index position laid $i$-th vector. Mathematically, this crossover can be expressed as follows:
\begin{equation}
%\vspace{-2mm}
\label{eq:de_xover}
w_{i,j}^{(t)}=\begin{cases}
u_{i,j}^{(t)}, & \text{rand}_{j}(0,1) \leq \mathit{CR} \vee j=j_{rand}, \\
x_{i,j}^{(t)}, & \text{otherwise},
\end{cases}
%\vspace{-2mm}
\end{equation}
\noindent where $\mathit{CR} \in [0.0,1.0]$ controls the fraction of parameters that are copied to the trial solution. The condition $j=j_{rand}$ ensures that the trial vector differs from the original solution $\mathbf{x}_{i}^{(t)}$ in at least one element. 

Mutation and crossover can be performed in several ways in DE. Consequently, in general, a specific notation was introduced to describe the varieties of these methods (also mutation strategies). For example, 'rand/1/bin' denotes that the base vector is selected randomly, 1 vector difference is added to it, and the number of modified parameters in the trial/offspring vector follows a binomial distribution. 

Mathematically, the selection can be expressed as follows:
\begin{equation}
\vspace{-2mm}
\label{eq:de_sel}
 \mathbf{x}_{i}^{(t+1)}=\begin{cases}
          \mathbf{w}_{i}^{(t)}, &\text{if } f(\mathbf{w}_{i}^{(t)}) \leq f(\mathbf{x}_{i}^{(t)}), \\
		  \mathbf{x}_{i}^{(t)}, &\text{otherwise}\,.
        \end{cases}
%\vspace{-2mm}
\end{equation}
The selection is usually called 'one-to-one', because the trial and the corresponding vector laid on the $i$-th position in the population compete for surviving in the next generation, where the better will survive according to the fitness function.

\subsection{Sankey diagrams}
Typically, Sankey diagrams are used for illustrating the quality and connectivity of flows between entities across time. Indeed, these diagrams present directed weighted graphs, where entities are represented as nodes connected by edges of different widths. The weights of the edges are referred to the quality of the flow. 

\begin{figure}[!htb]
    \centering
    \includegraphics[width=.96\linewidth]{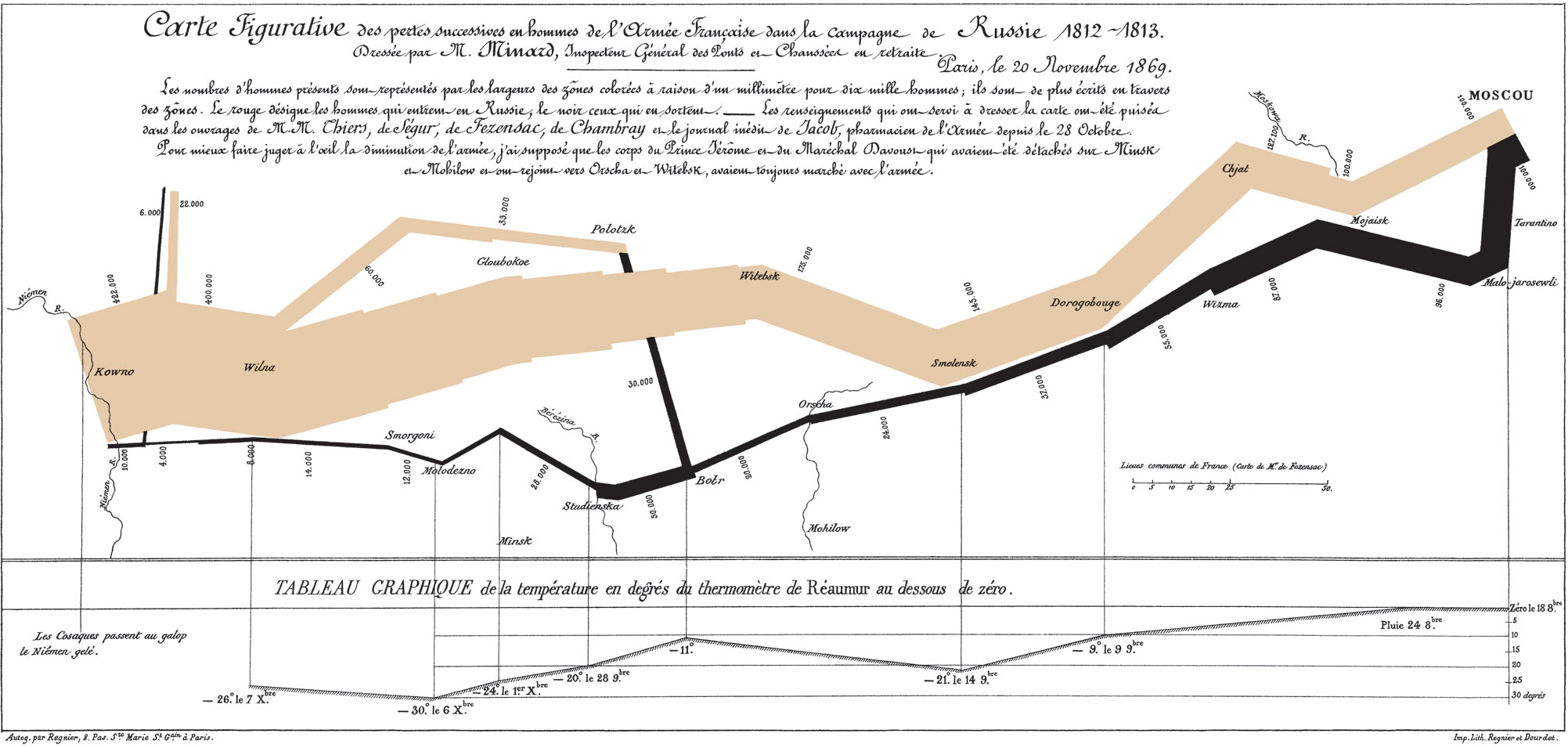}
    \caption{Minard's classic diagram of Napoleon's invasion of Russia, a predecessor of the Sankey diagram from~\cite{wikipediaEN2020classic}.}
    \label{fig:Minard}
\end{figure}

The pioneer work of this kind of visualization was made by Charles Minard, who presented Napoleon's Russian Campaign of 1812 using a flow diagram, denoting the amount of French soldiers, overlaid onto a geographical map (Fig.~\ref{fig:Minard}). 
These kind of diagrams were named after Captain Matthew Sankey, who used this method for visualizing energy transfer in steam engines~\cite{sankey1898introductory}.

The main problem in creating Sankey diagrams is the placement of nodes within layers such that crossings are avoided between edges (flows). Compared to the classical layered graph drawing problems, the Sankey creation is harder, due to the fact that each edge has a width connected to the value it encodes, and, therefore, crossings are of different importance. In general, this problem is defined as combinatorial and solved using various deterministic optimization algorithms, like the heuristic method proposed by Sugiyama et al.~\cite{sugiyama1981methods}, Integer Programming by Zarate et al.~\cite{zarate2018optimal}, and even a visual approach to understand and analyze the flow of information in a ML system by Chaudhuri et al.~\cite{chaudhuri2018visual}.

\subsection{Formal definition of the objectives}
The aim of visualization using Sankey diagrams for ARM is to search for those association rules that are distinguished by their similarities (quality aspect), and expose the good fitness (quantity aspect) regarding more time periods (temporal aspect). The spatial aspect of the map is reflected in the topology of nodes (attributes) connected with edges (attribute flows). Indeed, there are two issues that need to be considered for construction of the Sankey diagrams for ARM:
\begin{itemize}
    \item the map size,
    \item the similarity of the visualized association rules.
\end{itemize}
In summary, the Sankey diagram for ARM can be defined as a pair $\mathcal{F}=\langle M,R\rangle$, where $M$ limits the map size as:
\begin{equation}
    |R|\leq M,
\end{equation}
and the $R$ denotes the set of the most similar rules, determined according to the following method: Let us suppose that the results of the algorithm for ARM is a huge archive of association rules in the form of $X\Rightarrow Y$, which can, in general, be expressed as:
\begin{equation}
    X_1\wedge X_2\wedge \ldots \wedge X_n\Rightarrow Y_1\wedge Y_2\wedge \ldots \wedge Y_m,
    \label{eq:rule}
\end{equation}
where $X_i$ for $i=1,\ldots,n$ denotes the attributes of antecedent, and $Y_j$ for $j=1,\ldots,m$ the attributes of consequent sets, respectively. 

In order to divide both parts of the particular association rule, the following functions are introduced:
\begin{equation}
\begin{aligned}
\mathit{Ante}(X\Rightarrow Y)&=\{X_1,X_2,\ldots,X_n\},\\
\mathit{Cons}(X\Rightarrow Y)&=\{Y_1,Y_2,\ldots,Y_m\},
\end{aligned}
\end{equation}
where the $\mathit{Ante}(X\Rightarrow Y)$ function denotes the antecedent and the $\mathit{Cons}(X\Rightarrow Y)$ the consequent set of attributes contributed in the particular association rule $X\Rightarrow Y$. 

Then, the similarity between two rules $R_1$ and $R_2$ can be calculated as follows:
\begin{equation}
    \mathit{Sim}(R_1,R_2)=\frac{|\mathit{Ante}(R_1)\cap \mathit{Ante}(R_2)|+|\mathit{Cons}(R_1)\cap \mathit{Cons}(R_2)|}{|\mathit{Ante}(R_1)\cup \mathit{Ante}(R_2)|+|\mathit{Cons}(R_1)\cup \mathit{Cons}(R_2)|}.
\end{equation}
Let us mention that the value of $\mathit{Sim}(R_1,R_2)\in[0,1]$, where 1 means the full similarity, and 0 the full dissimilarity. Interestingly, the similarity values of the rules within the ARM archive can be combined into an adjacency matrix $\text{Adj}$, as follows:
\begin{equation}
    \text{Adj}=\left[ \begin{matrix}
    a_{1,1} & \ldots & a_{1,\tilde{N}} \\
     & \ldots &  \\
    a_{\tilde{N},1} & \ldots & a_{\tilde{N},\tilde{N}} \\
    \end{matrix} \right],
\end{equation}
where $a_{i,j}=\textit{sim}(R_i,R_j)$ for $i=1,\ldots,\tilde{N}$ and $j=1,\ldots,\tilde{N}$ presents a similarity value between rules $R_i$ and $R_j$, and $\tilde{N}\leq N$ denotes the number of the observed best association rules in the archive. 

Based on the introduced formalism, the problem of searching for the most similar set of association rules $R$ can be defined as a Knapsack 0/1~\cite{kellerer2004knapsack} problem in the following sense: Let us suppose $\tilde{N}$ objects and a knapsack of capacity $M$ are given. The problem is how to put into a knapsack the maximum number of objects of different weights $W=(w_1,\ldots,w_{\tilde{N}})$ such that the obtained profit $\text{Adj}=\{p_{i,j\}}$ is maximum. If a binary vector $B=(b_1,\ldots,b_{\tilde{N}})$ is adopted for denoting the inclusion of the object into the knapsack, the problem can be defined formally as follows:
\begin{equation}
    \max\sum_{i=1}^{\tilde{N}-1}a_{i,j}\cdot b_i,\qquad\text{for}~j=i+1,\ldots,\tilde{N},
    \label{eq:1}
\end{equation}
subject to:
\begin{equation}
    \max\sum_{i=1}^{\tilde{N}}w_{i}\cdot b_i,\leq M.
    \label{eq:2}
\end{equation}
Indeed, Eq.~(\ref{eq:1}) determines a lot of association rules with equal maximum values of similarity. 

\section{Method for creating the Sankey diagrams from the ARM archive} \label{sec:3}
\begin{figure}[!htb]
    \centering
    \includegraphics[width=.8\linewidth]{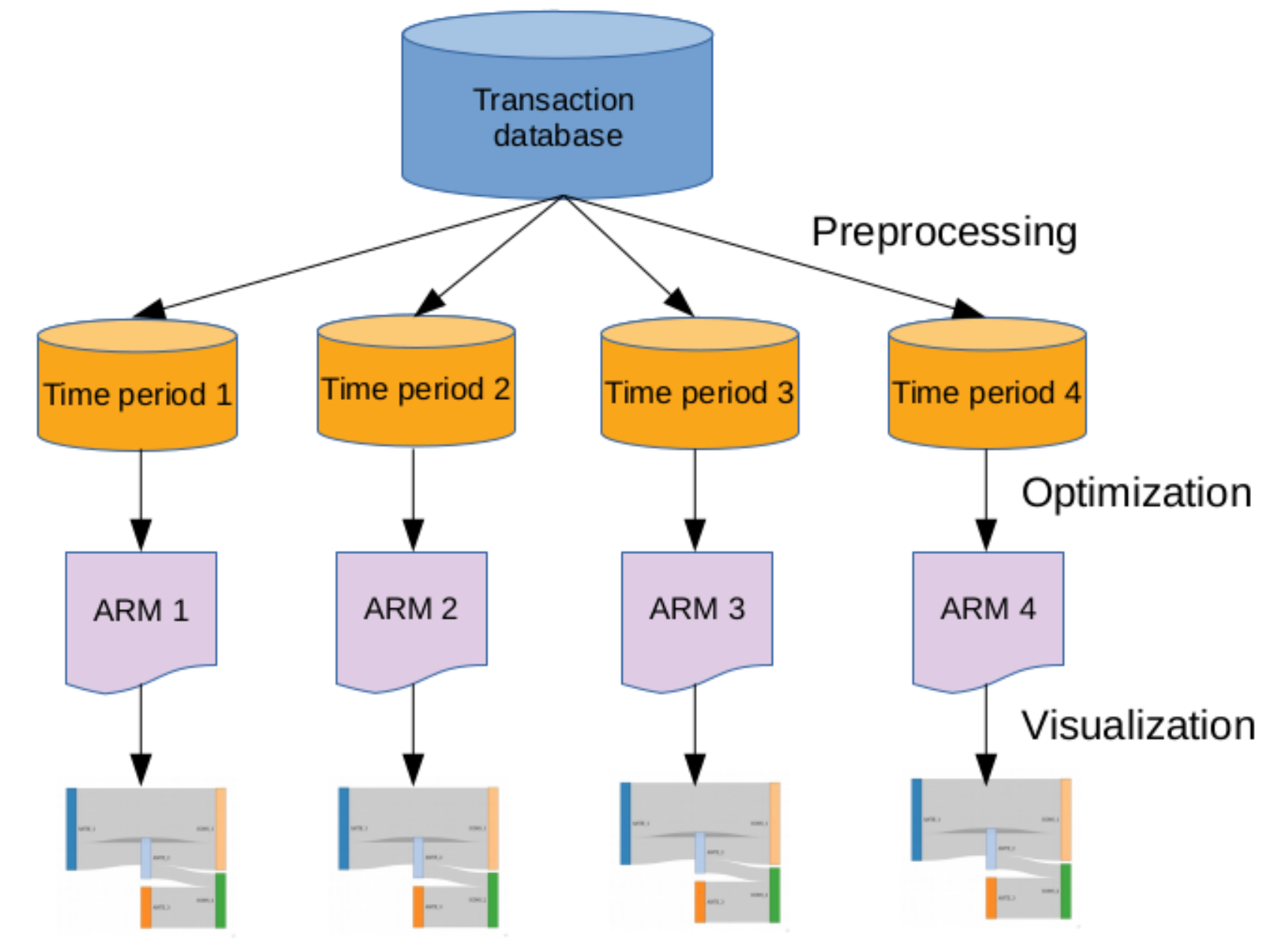}
    \caption{Architecture of the method for visualization of ARM using Sankey diagrams.}
    \label{fig:Sankey}
\end{figure}

A method for visualizing an archive of mined association rules using Sankey diagrams is divided into three steps (Fig.~\ref{fig:Sankey}):
\begin{itemize}
    \item preprocessing,
    \item optimization, 
    \item visualization.
\end{itemize}
In the remainder of the section, the mentioned steps are illustrated in detail.

\subsection{Preprocessing}
In the preprocessing step, a uniform transaction database is generally divided into $K$ auxiliary databases, where each of these captures transactions of the $K$-th time period. Normally, the performance improves from period to period. The motivation behind this selection in sport, for instance, lies in the fact that the best condition of the athlete must be turned on in the last time period, where the most important competitions are expected. The fact demands that the athlete needs to be prepared optimally at this time. On the other hand, the first time periods are dedicated to rest or performing sport training sessions of less intensity. 

However, the same method can also be applied for the other domains that are confronted with the association rules changing over time.

\subsection{Optimization}
Mining the association rules is performed in the proposed method using the DE algorithm for ARM, which demands modifying the following algorithm's components~\cite{eiben2015introduction}:
\begin{itemize}
    \item representation of individuals,
    \item genotype-phenotype mapping,
    \item fitness function evaluation.
\end{itemize}
Each solution in the proposed DE represents a mined association rule that is encoded of $d$-discrete attributes, to which the value for determining the ordering of the particular feature in the definite association rule is assigned. In addition, the last element of the vector denotes the so-called cut point, separating the antecedent from the consequent part of the association rule. As a result, the individual is represented as follows:
\begin{equation}
    \mathbf{x}^{(t)}_i=(\underbrace{\langle x^{(t)}_{i,0},x^{(t)}_{i,1} \rangle}_{\mathit{Feat}_1},\ldots,\underbrace{\langle x^{(t)}_{i,2(d-1)},x^{(t)}_{i,2d-1}\rangle}_{\mathit{Feat}_d},\underbrace{x^{(t)}_{i,2d}}_{\mathit{Cp}}),
\end{equation}
where each pair $\mathit{Feat}_j=\langle x^{(t)}_{i,2j},x^{(t)}_{i,2j+1}\rangle$ for $j=0,\ldots,d-1$ denotes the corresponding $j$-th feature with the first element designated decoded attribute, the second determining the ordering of this in a permutation $\Pi=(\pi_1,\ldots,\pi_d)$, and the $\mathit{Cp}$ is a cut point. Obviously, the variable $d$ indicates the maximum number of features, and the length of the individual is $2d+1$.

Genotype-phenotype mapping is used to map a representation of individuals in a genotype space to the variables in the problem space~\cite{eiben2015introduction}. In our case, the first variables in a pair $x^{(t)}_{i,2j}$, for $j=0,\ldots,d-1$, are decoded to the categorical attributes $\mathit{Attr}_j$ according to the equation:
\begin{equation}
    \mathit{Attr}_{\pi_j}=\left\lfloor \frac{x^{(t)}_{2j}}{|\mathit{Feat}_{\pi_j}|+1} \right\rfloor,\quad\text{for}~j=1,\ldots,d.
\end{equation}
Let us mention that $\mathit{Attr}_{\pi_j}=0$ has a special meaning, because it determines that the corresponding feature is not presented in the association rule.

A permutation of attributes in the association rule is calculated as follows: Each value $x^{(t)}_{2j+1}$ for $j=0,\ldots,d-1$ are assembled into an auxiliary vector $X'=(x^{(t)}_{2},\ldots,x^{(t)}_{2d-1})$. After ordering the descending vector, the permutation $\Pi=(\pi_1,\ldots,\pi_d)$ is constructed, such that the following relation is valid:
\begin{equation}
    x^{(t)}_{i,\pi_1}\leq x^{(t)}_{i,\pi_3}\leq\ldots\leq x^{(t)}_{i,\pi_{2d-1}}.
\end{equation}

The cut point is calculated according to the following equation:
\begin{equation}
    \mathit{Cp}_i=\lfloor x^{(t)}_{2d}\cdot(d-2)\rfloor+1, \quad\text{for}~i=1,\ldots,\mathit{Np},
\end{equation}
and delimits the antecedent from the consequent part of the particular association rule.

Fitness function is expressed as follows:
\begin{equation}
    f(\mathbf{x}^{(t)}_i)=\frac{\alpha\cdot\mathit{conf}(\mathbf{x}^{(t)}_i)+\beta\cdot\mathit{supp}(\mathbf{x}^{(t)}_i)}{\alpha+\beta},
    \label{eq:fit}
\end{equation}
where $\alpha$ and $\beta$ denote weights, while $\mathit{conf}(\mathbf{x}^{(t)}_i)$ and $\mathit{supp}(\mathbf{x}^{(t)}_i)$ are the confidence and the support of the mined association rule, respectively. 

\subsection{Visualization}
As can be seen from Eq.~(\ref{eq:rule}), mined association rules in the form $X\Rightarrow Y$express  explicitly the flow of a conjunction of antecedent attributes to a conjunction of consequent rules. Therefore, it is natural to represent them in a weighted graph, where the antecedent attributes denote the source and the consequent ones the sink nodes in the graph. Thus, the source nodes are joined with edges into a chain flowing to the specific sink nodes. The quality aspect of the graph is reflected by representing in the graph only the best $n$-association rules according to the similarity measure. 

When a weight is assigned to each edge, connecting source and drain nodes, that assesses the strength of the selected association rule in the sense of its fitness value, and the best mined association rules (i.e., flows) are compared in different time periods, also the temporary aspect of realizing the sport training sessions is obtained during the season. Thus, all preconditions for creating the Sankey diagram are satisfied.

Finally, a dynamic Knapsack 0/1 algorithm~\cite{horowitz1978fundamentals} is applied for searching the $M$ the most similar association rules appropriate for creating the Sankey diagrams. In order to prefer the more important ones, these rules are additionally distinguished according to their fitness function values.

\section{Experiments and results} \label{sec:4}
The goal of the experimental work was to show that the proposed method for association rules over time can be explained successfully by visualizing with Sankey diagrams. In line with this, improving the performance by an amateur cyclist during past seasons was analysed based on data obtained by realizing sport sessions using a wearable device.

The proposed method employs two algorithms: (1) The DE for ARM, and (2) The Dynamic KNAPsack 0/1 (DKNAP 0/1). The former is responsible for mining association rules, while the latter for explaining the content of huge ARM archives to the user by constructing the Sankey diagrams. The parameters of the mentioned algorithms used during the experimental work are illustrated in Table~\ref{tab:1}.  
\begin{table}[htb]
    \centering
    \caption{Parameter settings of the algorithms in tests.}
    \label{tab:1}
    \begin{tabular}{C{2cm}|L{5cm}|C{2cm}|R{2cm}}
    \hline
    Algorithm & Parameter & Abbreviation & Value  \\\hline\hline
\multirow{3}{*}{DE}	&	 Scale factor & $F$ & 0.5 \\
	&	 Crossover rate & $\mathit{CR}$ & 0.9 \\
	&	 Population size & $\mathit{Np}$ & 100 \\ \hline
\multirow{2}{*}{DKNAP 0/1}	&	Map size & $M$ & 4 \\ 
	&	Number of observed rules & $N$ & 100 \\ \hline
    \end{tabular}
\end{table}

Let us mention that the best results of the DE algorithm, obtained after 25 independent runs, were considered for visualization. Obviously, these results were assessed according to the fitness function value expressed by Eq.~(\ref{eq:fit}). On the other hand, the result of the DKNAP 0/1 was obtained after one run.

\subsection{Transaction database}
The mobile data obtained by wearable devices were saved into a transaction database that was divided into $K=4$ auxiliary databases according to the different time periods. In our case, the first time period captures transactions realized from March 2013 to May 2014, the second time period from May 2014 to July 2015, the third time period from July 2015 to May 2016, and the last time period from October 2016 to October 2017 (Table~\ref{tab:2}). Interestingly, the sizes of the auxiliary databases can also be seen from the Table.

\begin{table}[htb]
    \centering
    \caption{Characteristics of auxiliary databases.}
    \label{tab:2}
    \begin{tabular}{C{2cm}|L{4cm}|R{5cm}}
    \hline
    Part & Time period & Number of transactions  \\\hline\hline
1	&	 March 2013 to May 2014 & 126 \\
2	&	 May 2014 to July 2015 & 134 \\
3	&	 July 2015 to May 2016 & 139 \\
4	&	 October 2016 to October 2017 & 136 \\ \hline
Total	&	 The whole dataset & 535 \\\hline
    \end{tabular}
\end{table}

As can be seen from Table~\ref{tab:3}, each transaction is described by seven discrete features indicating the performance of the athlete during realization of a particular training session. To each feature in the table, the corresponding domain of values is attached that is represented as a discrete set of feasible attributes. 

\begin{table}[htb]
    \centering
    \caption{Observed discrete features with their corresponding attributes.}
    \label{tab:3}
    \begin{tabular}{C{1cm}|L{3cm}|L{2.2cm}|R{5cm}}
    \hline
    Nr. & Feature & Abbreviation & Attributes \\\hline\hline
1	&	 Duration & DURATION & \{SHORT,MEDIUM,LONG\} \\
2	&	 Distance & DISTANCE & \{SHORT,MEDIUM,LONG\} \\
4	&	 Calories & CALORIES & \{SMALL,MEDIUM,HIGH\} \\
3	&	 Average Heart Rate & HR & \{MEDIUM,HIGH\} \\
5	&	 Altitude & ALTITUDE & \{LOW,MEDIUM,HIGH\} \\
6	&	 Ascent & ASCENT & \{LOW,MEDIUM,HIGH\} \\
7	&	 Descent & DESCENT & \{LOW,MEDIUM,HIGH\} \\\hline
    \end{tabular}
\end{table}

Let us mention that attributes in the mined association rules are represented as a concatenation of a feature name by an attribute name using character '\_' for joining both. Obviously, the DE for ARM is applied on each of the auxiliary databases independently. As a result, four independent archives of mined association rules are obtained that further enters into the visualization process. 

\subsection{The results}
The results of the proposed method for ARM changing over time need to be analyzed after two steps: The first step highlights the results of an optimization, while the second a visualization of the obtained data using Sankey diagrams. The results of the optimization are illustrated in Table~\ref{tab:4} that presents the four best association rules according to the fitness function value. 
\begin{table}[htb]
    \centering
    \caption{The best association rules selected for visualization.}
    \label{tab:4}
    \small
    \begin{tabular}{C{.6cm}|C{1.2cm}|L{8.2cm}|R{1.2cm}}
    \hline
    Part & Rule nr. & Association rule & Fitness\\\hline\hline
\multirow{4}{*}{1}	& 1 & CALORIES\_SMALL $\Rightarrow$ ASCENT\_LOW & 0.7262\\
	& 2 & DURATION\_SHORT $\Rightarrow$ ASCENT\_LOW & 0.6661\\
	& 3 & CALORIES\_SMALL $\wedge$ DURATION\_SHORT $\Rightarrow$ ASCENT\_LOW & 0.6439\\
	& 4 & CALORIES\_SMALL $\wedge$ DISTANCE\_SHORT $\Rightarrow$ ASCENT\_LOW & 0.6293\\ \hline
\multirow{4}{*}{2}	& 1 & CALORIES\_SMALL $\Rightarrow$ ASCENT\_LOW & 0.7612\\
	& 2 & DURATION\_SHORT $\Rightarrow$ ASCENT\_LOW & 0.7015\\
	& 3 & CALORIES\_SMALL $\wedge$ DURATION\_SHORT $\Rightarrow$ ASCENT\_LOW & 0.6977\\
	& 4 & DISTANCE\_SHORT $\Rightarrow$ ASCENT\_LOW & 0.6903\\\hline 
\multirow{4}{*}{3}	& 1 & CALORIES\_SMALL $\Rightarrow$ ASCENT\_LOW & 0.7446\\
	& 2 & DISTANCE\_SHORT $\Rightarrow$ ASCENT\_LOW & 0.7122\\
	& 3 & DURATION\_SHORT $\Rightarrow$ ASCENT\_LOW & 0.7086\\ 
	& 4 & CALORIES\_SMALL $\wedge$ DURATION\_SHORT $\Rightarrow$ ASCENT\_LOW & 0.7014\\\hline
\multirow{4}{*}{4}	& 1 & DURATION\_SHORT $\Rightarrow$ ASCENT\_LOW & 0.7500\\
	& 2 & DISTANCE\_SHORT $\Rightarrow$ ASCENT\_LOW & 0.6875\\
	& 3 & HR\_HIGH $\Rightarrow$ ASCENT\_LOW & 0.6744\\
	& 4 & DURATION\_SHORT $\wedge$ DISTANCE\_SHORT $\Rightarrow$ ASCENT\_LOW & 0.6691\\\hline
    \end{tabular}
    \normalsize
\end{table}
As can be seen from the Table, relatively simple association rules are preferred by the DE for ARM, i.e., one consequent follows from one antecedent. On the other hand, all consequents in the Table converge to the same attribute, ASCENT\_LOW. As antecedents in compound association rules, typically, a permutation of two attributes are taken from a set 
\{CALORIES\_SMALL, DURATION\_SHORT,DISTANCE\_SHORT\}.

\begin{figure*}[hbt]
\centering
\subfloat[Time period 1]{
\begin{minipage}{0.49\textwidth}
\includegraphics[width=0.99\textwidth]{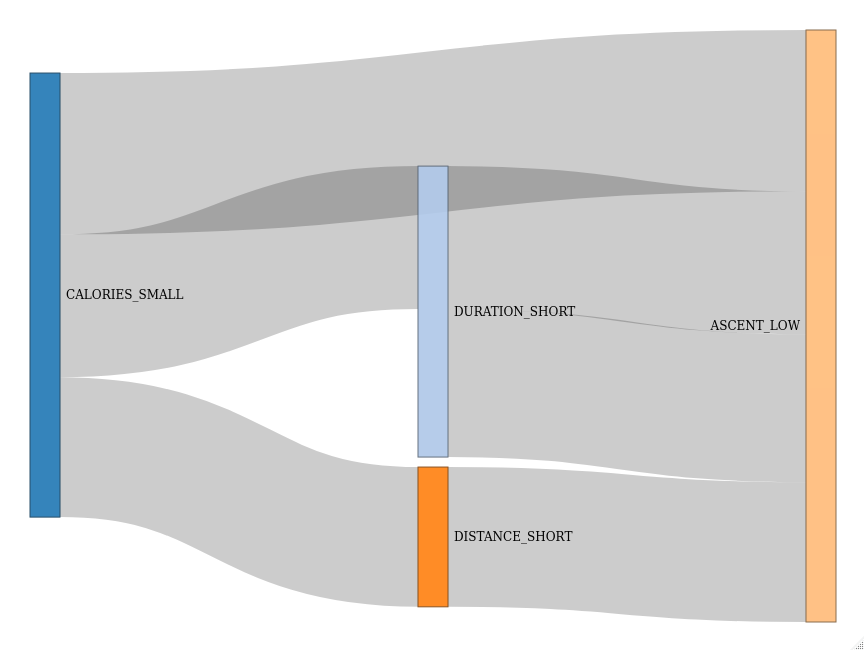}
\end{minipage}}
\subfloat[Time period 2]{
\begin{minipage}{0.49\textwidth}
\includegraphics[width=0.99\textwidth]{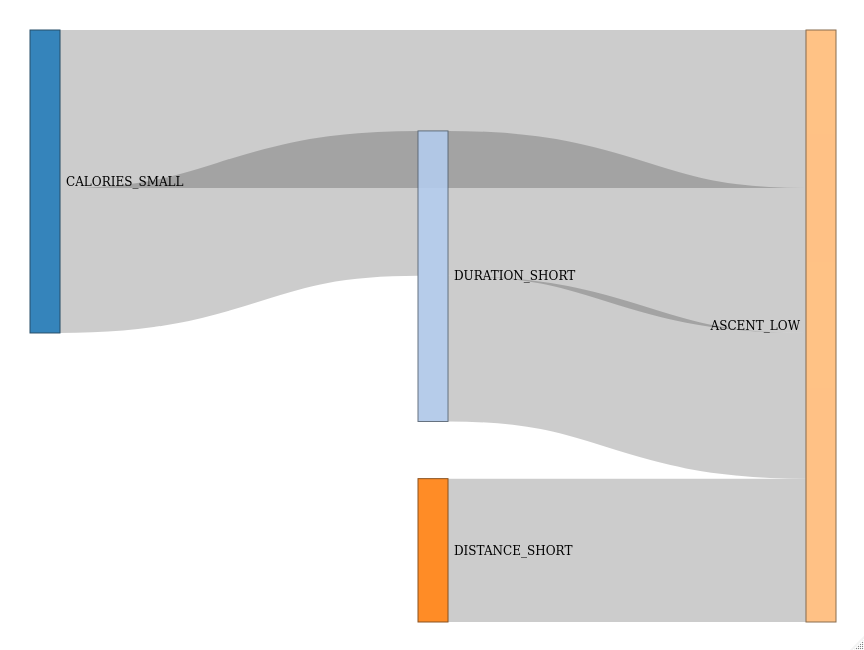}
\end{minipage}}
\qquad
\subfloat[Time period 3]{
\begin{minipage}{0.49\textwidth}
\includegraphics[width=0.99\textwidth]{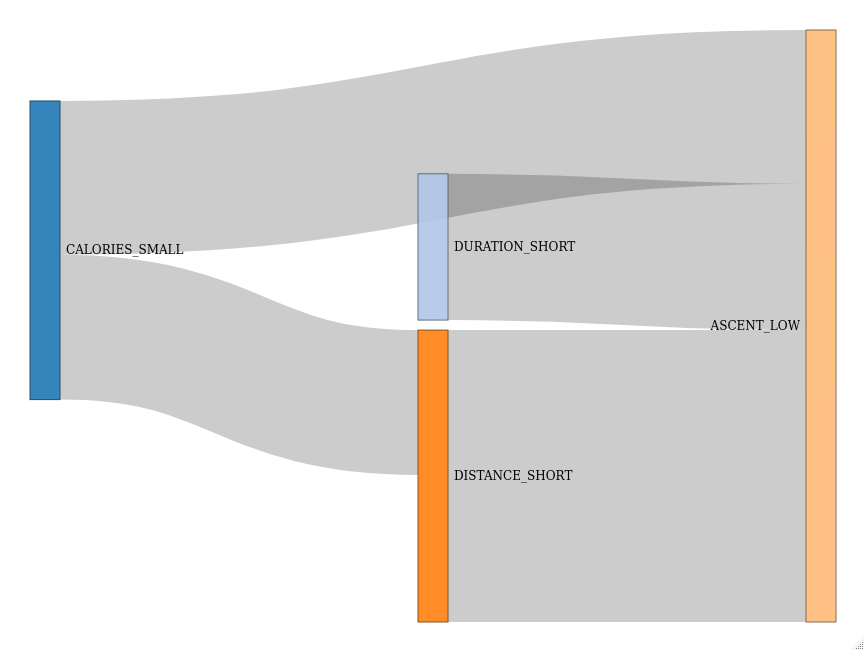}
\end{minipage}}
\subfloat[Time period 4]{
\begin{minipage}{0.49\textwidth}
\includegraphics[width=0.99\textwidth]{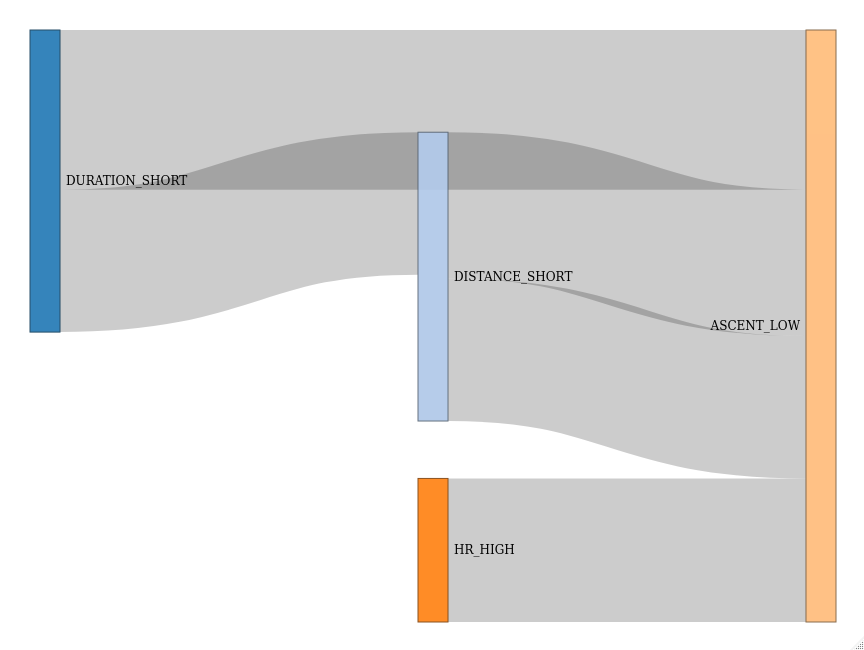}
\end{minipage}}
\caption{Visualization of the most similar four association rules using Sankey diagrams.}
\label{fig:scenario-1}
\end{figure*}

Visualization of the results presented in Table~\ref{tab:4} are depicted in Fig.~\ref{fig:scenario-1} that is divided into four diagrams reflecting the performance of an athlete in four different time periods. Diagrams were visualized using the network3D package~\cite{allaire2017networkd3}.
As can be seen from the Figure, the CALORIES\_SMALL attribute is the most important antecedent in the first time period, due to producing the widest flow in the corresponding Sankey diagrams. Indeed, this flow diminished till the end of the observed data, where it disappeared. In place of this attribute, an antecedent DURATION\_SHORT and DISTANCE\_SHORT prevailed. Interestingly, the antecedent HR\_HIGH emerged in the last time period, denoting that the feature HR is a more important factor for determining the performance of an athlete than the CALORIES. 

\subsection{Discussion}
Visualization of association rules changing over time with Sankey diagrams showed many advantages that can be summarized as follows: These diagrams are appropriate for representing the flow (topology) of a set of antecedents to different consequents. Actually, the antecedents are a sequence of source nodes, while the consequents represent sink nodes in the Sankey diagrams. 

A temporary aspect of the presentation is accomplished by maintaining four auxiliary databases, within which searching for performances of the athlete is conducted temporally, and the results are depicted in Sankey diagrams. From the sequence of these diagrams, a sport trainer can conclude, which relations of attributes are more important for an athlete in a specific time period.

\section{Conclusion} \label{sec:5}
Sophisticated AI powered systems operate nowadays without any human intervention for design and deployment on the one hand, but they have become hard to understand for the users on the other. Therefore, the new domain of EAI has emerged that tries to avoid these limitations of the current AI systems. The EAI searches for answers to the following two issues: (1) How to create more explainable ML models, and (2) How to enable users to understand, trust, and manage the emerging AI platforms effectively. More post-hoc techniques have emerged for dealing with the first issue, where a visual explanation is one of the more prominent ones.

The aim of the chapter is to propose the new explainable method for visualization mined association rules over time. The method consists of three steps: preprocessing, optimization, and visualization. The first step is devoted to dividing the uniform transaction database into more auxiliary databases corresponding to definite time periods. In the second step, the archives of association rules mined by DE are discovered from each auxiliary database. These are then visualized using Sankey diagrams, which are capable of representing multivariate data from different aspects, i.e., quality, quantity, topological, and temporal. In these diagrams, each attribute in association rules denotes a node in the corresponding direct graph, while edges indicate the relations between these. In addition, the edges are of different widths that are proportional to the fitness of the association rule. 

The proposed method was applied to the transaction database created by an amateur cyclist during the past seasons. Each training session was monitored by a mobile device worn by an athlete during the realization, and saved as a transaction into a transaction database. The transaction is characterized by discrete attributes. The transaction database was divided into four auxiliary databases each denoting one fourth of the past seasons. DE for ARM was employed to each of the auxiliary databases, producing four archives of the association rules. Four of the best association rules according to similarity and fitness are entered into the visualization using Sankey diagrams. The result of visualization showed that a trend in improving performance of an athlete can be indicated by changing the attributes appearing in the selected association rules in different time periods.

The preliminary study offers many directions for the future development: At first, the method could be applied to other transaction databases (e.g., UCI Machine Learning Repository). Then, DE could be replaced by another stochastic nature-inspired population based algorithms. Finally, another kind of calculating the similarity between association rules could be developed.

\end{document}